%% file: main.tex
\documentclass{article}
\pdfoutput=1
\usepackage{iclr2027_conference,times}
\iclrfinalcopy

\input{math_commands.tex}

\usepackage{hyperref}
\usepackage{url}
\usepackage{amsmath}
\usepackage{amssymb}
\usepackage{booktabs}
\usepackage{graphicx}
\graphicspath{{figures/}}
\usepackage{multirow}
\usepackage{array}
\usepackage{bm}
\usepackage{xcolor}
\definecolor{deltagreen}{rgb}{0.0,0.45,0.0}
\definecolor{deltared}{rgb}{0.6,0.0,0.0}
\newcommand{\dgain}[1]{\,{\scriptsize\textcolor{deltagreen}{(+#1)}}}
\newcommand{\dloss}[1]{\,{\scriptsize\textcolor{deltared}{($-$#1)}}}

\newcommand{\dfaster}[1]{\,{\scriptsize\textcolor{deltagreen}{($-$#1)}}}
\newcommand{\dslower}[1]{\,{\scriptsize\textcolor{deltared}{(+#1)}}}

\usepackage{soul}
\usepackage{float}
\usepackage{placeins} %
\usepackage{etoolbox}
\AtBeginEnvironment{figure}{\setlength{\abovecaptionskip}{-6.97pt}}
\definecolor{hlmark}{HTML}{FFE9B3} %
\sethlcolor{hlmark}
\newcommand{\shared}[1]{\hl{#1}}
\newcommand{\sharedn}[2]{\hl{#2}\textsuperscript{\scriptsize #1}}
\definecolor{palgreen}{HTML}{009E73}
\definecolor{palblue}{HTML}{0072B2}
\definecolor{palamber}{HTML}{E49A4A}
\colorlet{hlc1}{palgreen!28}
\colorlet{hlc2}{palblue!22}
\colorlet{hlc3}{palamber!40}
\newcommand{\sharedc}[2]{{\sethlcolor{hlc#1}\hl{#2}}}
\newcommand{\sharedcn}[2]{\sharedc{#1}{#2}\textsuperscript{\scriptsize #1}}
\newlength{\exlabw}
\newcolumntype{Y}{m{\exlabw}}          %
\newcolumntype{Z}{m{\dimexpr\linewidth-\exlabw-2\tabcolsep\relax}} %
\newcolumntype{W}{m{\dimexpr(\linewidth-\exlabw-4\tabcolsep)/2\relax}} %
\newcommand{\exlab}[1]{\hspace*{4pt}\textbf{#1}}
\newcommand{\exitem}[2]{\ifnum#1>1\par\addvspace{2pt}\fi\hangindent=1.4em\hangafter=1 #1.~#2}

\title{Online Label-Free Gradient Matching for \\ Continual Language Model Training}

\author{Sajjad Pakdamansavoji$^{1,*}$, Mohammad Reza Rezaei$^{1}$, Ilija Subasic$^{1}$, Pedro Garzon$^{1}$, \\
Radu Paradovschi, Zhao Chen$^{1}$ \\[4pt]
{\mdseries $^{1}$Upwork Inc (\url{upwork.com}), $^{*}$Corresponding author}
}

\makeatletter
\def\@maketitle{\vbox{\hsize\textwidth\centering
{\LARGE\sc \@title\par}
{\bf\rule{\z@}{24pt}\@author\par}
\vskip 0.3in minus 0.1in}}
\makeatother

\newcommand{\thet}{\bm{\theta}}
\newcommand{\vct}[1]{\bm{#1}}

\newcommand{\Dreal}{\gD_{\text{real}}}
\newcommand{\Dsyn}{\gD_{\text{syn}}}

\DeclareMathOperator{\PMI}{PMI}

\begin{document}

\maketitle
\lhead{Preprint}

\begin{abstract}
Language models in customer-facing products must be continually updated, since the data they serve shifts over time. As retraining on the full accumulated data is impractical, each update should instead run on a small set of examples carrying the training signal of all data seen so far. The strongest current methods build this set by gradient matching, which synthesizes examples whose training gradients match those of the full data. However, they require ground-truth labels, run once on a fixed snapshot rather than on a growing stream, and are slow on large data. We introduce \textbf{OLGA}, which lifts these limitations. It distills unlabeled data by drawing the gradient matching targets from the model's own next-token predictions. Across five text classification benchmarks with phi-1.5, it improves accuracy by 10 percentage points over the strongest selection baseline using as few as five examples. OLGA also distills online data streams by keeping a running rehearsal memory of synthetic examples, which is re-distilled together with each new batch as it arrives. In the online setting, it outperforms the strongest selection-based continual baseline by 6.6 stream-AUC percentage points. Finally, our method scales to large data by pairing distillation with selection in a hybrid pipeline, letting selection handle the mild stage of compression and distillation the extreme one. It cuts runtime by 30\% versus distillation alone at similar accuracy, and beats the best selection-only pipeline by up to 12.6 percentage points.
\end{abstract}

\section{Introduction}
\label{sec:intro}

Language models are increasingly deployed in customer-facing products, where the traffic they serve keeps changing. User preferences move, new topics appear, and safety requirements shift, so a model that was accurate at release gradually falls out of step with the data it now sees~\citep{lazaridou2021mind}. Keeping it current means updating it repeatedly rather than once, and the data available at each update grows with every cycle. Retraining on every example accumulated so far is the obvious remedy and also the least affordable one, because a full pass costs more as the data grows while the oldest examples contribute least. What a practical update loop needs instead is a small set of examples, rebuilt at each cycle, that still carries the training signal of everything seen so far.

Most existing methods build that set by selecting a well-chosen subset of the real examples. This works well when the budget is generous, but once only a few examples can be kept, it becomes hard to preserve a strong training signal. Dataset distillation with gradient matching addresses this by synthesizing examples whose training gradient matches the gradient of the real data~\citep{zhao2021dataset}. Because a synthetic example does not have to be a real sentence, it can capture what many real examples teach the model, so the set can be kept to a few examples.

Three obstacles stand in the way of using gradient matching to keep a language model current. First, it needs ground-truth labels, which rules out the unlabeled data that makes up most of what a deployed system collects. For images, recent work replaces the labels with targets from a frozen model trained without them~\citep{lee2024self,joshi2025dataset}, but text distillation methods still rely on labels~\citep{ogawa2025technical}. Second, it runs once over a fixed snapshot, so it never has to fold an arriving batch into an existing set. Methods that do this exist for images~\citep{gu2024summarizing}, but they need labels and nudge pixel values in small steps, which does not work for the discrete tokens of text. Finally, the cost of matching gradients grows with the amount of data it compresses, making it least affordable exactly when data is plentiful. For images, discarding part of the real data before distilling cuts this cost~\citep{xu2024distill,moser2025distill}, but these methods rank examples by their loss against a label and have only been studied on a fixed snapshot. As a result, no existing method compresses an unlabeled text stream into a small training set at a cost that stays affordable as the stream grows.

We remove all three obstacles at once. To drop the labels, we keep them out of the objective and take the matching target from the model's own predictions instead. To move from a snapshot to a stream, we keep a running rehearsal memory of synthetic examples and merge each arriving batch into it with the same gradient matching objective. To control the cost, we introduce a hybrid pipeline in which cheap unsupervised selection runs ahead of expensive distillation, since the two are strong at different amounts of compression.

We make the following contributions.

\begin{itemize}
  \item We introduce a label-free gradient matching objective whose compact distilled set trains a model 10 points more accurate than the best real subset of the same size.
  \item We extend text distillation to streaming data, updating a rehearsal memory as each batch arrives, and beat the best real memory of the same size by 6.6 stream-AUC points.
  \item We design a hybrid pipeline in which cheap selection shrinks each batch before distillation, cutting distillation runtime by 30\% compared with distilling the full batch.
\end{itemize}

\section{Background and Related Work}
\label{sec:related}

For much of the past decade, progress in deep learning came from scale, with larger models trained on more data consistently outperforming smaller ones. As models have grown to billions of parameters and corpora to trillions of tokens, however, the bottleneck has shifted from data availability to compute and time. Deployed models must also be updated ever more often as user preferences, knowledge, and safety requirements change, and retraining on the full corpus at every update is prohibitively expensive. The question is therefore no longer how to collect more data, but how to select or construct the smallest set of examples that preserves the training signal of the full corpus.

\paragraph{Coreset selection.}
Coreset selection keeps the most representative slice of the real training set without modifying any examples. Geometry-based algorithms formalize this as a coverage problem, where k-center greedy~\citep{sener2018active} picks the point farthest from the selected set, while herding~\citep{rebuffi2017icarl} and prototypical selection~\citep{snell2017prototypical} sequentially choose samples that best approximate the per-class feature centroid. Training-oriented variants use model-dependent signals such as gradient norms (GraNd) or L2 prediction error (EL2N)~\citep{paul2021datadiet} to prioritize samples near the decision boundary rather than near the class mean. Despite this diversity, all coreset methods share a hard representational ceiling, since a text sequence is an atomic unit that cannot be perturbed below the token level and no selection rule can manufacture variance that does not already exist in the chosen sample. This ceiling becomes binding at extreme compression, the regime where synthetic examples become necessary.

\paragraph{Dataset distillation.}
Dataset distillation compresses a large training set into a handful of synthetic examples, optimized so that a model trained on them behaves like one trained on the original data. \citet{wang2018dataset} first framed this as bi-level meta-learning, updating the synthetic examples through an unrolled inner training loop, but the memory cost and numerical instability of unrolling restrict it to shallow networks trained for only a few steps. Later work replaces the unroll with cheaper surrogate objectives. \emph{Trajectory matching}~\citep{cazenavette2022dataset} aligns the parameter path traced by training on synthetic data with paths recorded on real data, at the cost of pre-computing and caching thousands of such expert trajectories. \emph{Distribution matching}~\citep{zhao2023dataset} instead matches per-class feature embeddings of real and synthetic examples, which is cheap but ignores the task loss and can leave class boundaries misaligned. \emph{Gradient matching}~\citep{zhao2021dataset} targets the task loss directly by finding synthetic examples whose per-step parameter gradient matches the real gradient in cosine distance. It needs neither long unrolls nor stored trajectories and scales to larger architectures more readily, making it the dominant paradigm. All three directions, however, assume that gradients flow through a continuous input space, which holds for images but breaks down for discrete text tokens, where a gradient step on the input has no direct meaning.

\paragraph{Dataset distillation for language models.}
Early work on distilling text avoided discreteness by optimizing continuous word embeddings instead of tokens~\citep{maekawa2023attention}, but the results are model-specific, meaning a synthetic embedding created with BERT cannot be used to fine-tune RoBERTa, and projecting the optimized embeddings back to the nearest vocabulary token yields incoherent sequences that carry little training signal. \citet{maekawa2024dilm} addressed this with DiLM, which trains a proxy generative model to synthesize readable text, routing gradients through the discrete bottleneck via generation probabilities and producing sequences that transfer across architectures. Training a full proxy language model as a distillation scaffold is itself expensive, however, and the proxy may memorize sensitive training-set content. GradMM~\citep{nguyen2025synthetic} avoids the proxy by applying the Alternating Direction Method of Multipliers (ADMM)~\citep{boyd2011distributed}, in which gradient matching proceeds in continuous embedding space while a periodic projection step snaps each embedding back to its nearest vocabulary token, enforcing hard discreteness throughout optimization. This produces human-readable, cross-architecture synthetic text, and it is the offline supervised technique we build on here. Like every text distillation method so far, however, it places ground-truth labels in the matching objective, which ties the distilled data to a single task and excludes the unlabeled data that makes up most of what language models are trained on.

\paragraph{Continual learning and catastrophic forgetting.}
In continual learning, a model must keep absorbing a non-stationary stream of data without losing what it has already learned. The core difficulty is \emph{catastrophic interference}, in which training on new material abruptly overwrites earlier knowledge~\citep{mccloskey1989catastrophic,french1999catastrophic}, and it arises whether the stream introduces new tasks, new domains, or new classes~\citep{van2019three}. Architecture-based methods counter it by dedicating parameters to each task~\citep{rusu2016progressive,mallya2018packnet}, but presuppose known task boundaries and a growing parameter budget. Regularization-based methods penalize changes to weights that matter for past tasks~\citep{kirkpatrick2017overcoming,zenke2017continual,li2018learning}, but weaken over long streams as their accumulated constraints stiffen the network. The dominant remedy is \emph{rehearsal}, which replays a small memory of past examples alongside incoming data, with iCaRL~\citep{rebuffi2017icarl} as its canonical instance and a strong baseline for nearly a decade. Because the memory \emph{is} the method, much work concerns how to curate it, from fixed rules such as reservoir sampling, herding, or class-balanced storage~\citep{prabhu2020gdumb} to gradient-based rules that keep the examples whose gradients best span~\citep{aljundi2019gradient} or best approximate~\citep{tiwari2022gcr} the gradient of all data seen so far. Every such scheme still chooses among real examples, so the memory inherits the representational ceiling of coreset selection. That ceiling binds hardest where rehearsal matters most: as the stream grows and the per-class budget shrinks, a few stored examples can no longer carry the training signal of the task they stand for.

\paragraph{Continual distillation as rehearsal memory.}
This ceiling motivates compressing the memory rather than selecting it, using dataset distillation to synthesize a handful of maximally informative rehearsal examples. Distilled Replay keeps as little as one synthetic pattern per class~\citep{rosasco2022distilled}, \citet{zhao2021dataset} showed that gradient-matched synthetic memories alleviate forgetting in a task-incremental image setting, and later methods refine the buffer by carrying cross-task information~\citep{liao2025data}, mixing distilled with real exemplars~\citep{kong2024hybrid}, or condensing prototypical exemplars~\citep{nguyen2026memory}. Two limitations run through this line of work. First, these methods optimize continuous pixel tensors with objectives that do not carry over to discrete tokens. Continual learning for text has therefore sidestepped distillation in favor of generative rehearsal, sampling pseudo-examples from an autoencoder~\citep{zeng2024dirichlet} or prompting the model to synthesize its own rehearsal set~\citep{huang2024mitigating}, which re-incurs the expense and memorization risk of a separate generator. Second, nearly all of them are offline, distilling each task once from a complete snapshot. The one exception, SSD~\citep{gu2024summarizing}, folds each arriving batch into a synthetic image memory by matching gradients against the labeled batch. How to fold an arriving batch into a synthetic memory of discrete tokens, with or without labels, has not been studied.

\section{Methodology}
\label{sec:method}

\begin{figure}[t]
\begin{center}
\includegraphics[width=\columnwidth]{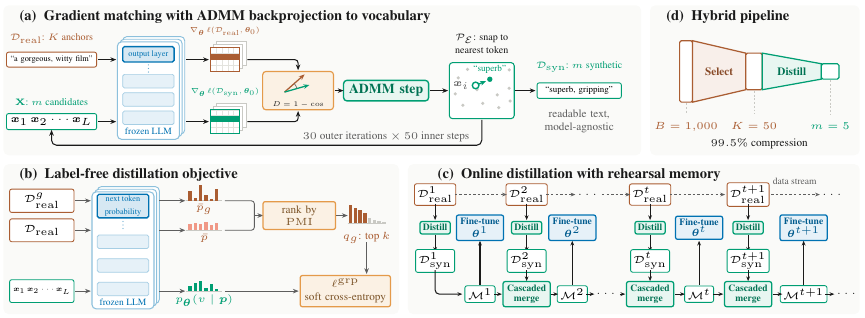}
\end{center}
\caption{Overview of the components of OLGA. \textbf{(a)} Synthetic text is generated by matching its training gradient to that of real data, alternating with a step that maps it back to real tokens. \textbf{(b)} A frozen language model's next-token predictions replace ground-truth labels, so gradient matching works on unlabeled data. \textbf{(c)} A running rehearsal memory extends offline distillation to an online data stream. Each new batch is distilled together with the memory, which the model then retrains on. \textbf{(d)} Our hybrid design first uses cheap selection to shrink each batch, then distills what remains into the few examples kept, which cuts runtime on large data.}
\label{fig:overview}
\vspace{-13pt}%
\end{figure}

To address these obstacles, we introduce \underline{O}nline \underline{L}abel-free \underline{G}radient M\underline{a}tching (\textbf{OLGA}), in which each component removes one obstacle (Figure~\ref{fig:overview}). We begin in Section~\ref{sec:method-gradmm} with the offline supervised method that OLGA builds on. Section~\ref{sec:method-labelfree} then drops the need for labels, Section~\ref{sec:method-online} extends the method to streaming data, and Section~\ref{sec:method-hybrid} speeds it up with a hybrid design.

\subsection{Gradient matching with the supervised objective}
\label{sec:method-gradmm}

The method we start from is GradMM~\citep{nguyen2025synthetic}, which produces readable and architecture-agnostic examples by solving a discrete gradient matching optimization. Let $\thet_0$ denote the parameters of a frozen pretrained language model and $\gD$ a set of examples $s=(\vct p,\vct r)$, each pairing a prompt with a response. The loss of the model on $\gD$, written $\ell(\gD,\thet)$, is the average of the individual example losses $\ell(s,\thet)$. The real examples that serve as a matching target are called \emph{anchors} and written $\Dreal$, and the synthetic set produced from them is $\Dsyn$, held to a budget of $m$ examples. GradMM searches for a synthetic set whose fine-tuning gradient matches the anchor gradient at the pretrained parameters,
\begin{equation}
\label{eq:gradmm}
\Dsyn^{\star}=\argmin_{\substack{|\Dsyn|\le m,\ \Dsyn\subset\Gamma}}\;
D\!\left(\nabla_{\thet}\,\ell(\Dsyn,\thet_0),\ \nabla_{\thet}\,\ell(\Dreal,\thet_0)\right),
\end{equation}
where $D(\vct u,\vct v)=1-\cos(\vct u,\vct v)$ compares gradient vectors by direction rather than scale, and $\Gamma$ is the set of readable token sequences. GradMM solves this with ADMM~\citep{boyd2011distributed}, which alternates a few steps of first-order optimization in continuous embedding space with a projection back onto real vocabulary tokens (Figure~\ref{fig:overview}(a)). We keep this solver unchanged and focus instead on making the optimization work without labels, on a data stream, and at lower runtime.

In the supervised setup, the model is asked to emit a verbalizer token $v_y$, a single word standing in for the class of the example, at a fixed position after the prompt, and the loss is the cross-entropy of that single token, $\ell^{\mathrm{sup}}(s,\thet)=-\log p_{\thet}(v_y \mid \vct p)$. Since every example of a class is trained toward the same token, their gradients align and add up rather than cancel. The anchor gradient thus lies in a few class directions, which a handful of synthetic examples can reproduce closely.

\subsection{Gradient matching with a label-free objective}
\label{sec:method-labelfree}

While the supervised loss $\ell^{\mathrm{sup}}(s,\thet)$ needs each example's class to know which verbalizer token the model should emit, customer-facing platforms mostly receive unannotated examples for which a single class label is hard to pin down~\citep{zhao2024wildchat}. Such data is rarely unstructured, though, and can be partitioned into groups $\Dreal^{g}$ by pre-logged metadata or through unsupervised clustering~\citep{wan2024tnt}. We use this grouping, together with the model's internal next-token probabilities, to design a label-free loss $\ell^{\mathrm{grp}}(s,\thet)$ that preserves inter-group separation (Figure~\ref{fig:overview}(b)).

Let $p_{\thet_0}(v\mid\vct p)$ denote the probability the frozen model puts on a next token $v$ after a prompt $\vct p$, and let $\bar p_{g}(v)$ and $\bar p(v)$ be its averages over one group and over all anchors. The pointwise mutual information between a token and a group,
\begin{equation}
\label{eq:pmi}
\PMI(v,g)=\log\bar p_{g}(v)-\log\bar p(v)
=\log\frac{\frac{1}{|\Dreal^{g}|}\sum_{s\in\Dreal^{g}}p_{\thet_0}(v\mid \vct p)}
           {\frac{1}{|\Dreal|}\sum_{s\in\Dreal}p_{\thet_0}(v\mid \vct p)},
\end{equation}
measures how much more the model expects $v$ inside one group than across the data as a whole. Ranking the vocabulary by \eqref{eq:pmi} and keeping the $k$ highest-scoring tokens gives a set $\gV_{g}^{k}$ of the tokens that best separate group $g$ from the rest, and $k$ is the only hyperparameter the construction adds. Normalizing the group probabilities across $\gV_{g}^{k}$ turns them into a target $q_{g}$, and for $s\in\Dreal^{g}$ the group-distribution loss is the soft cross-entropy toward it,
\begin{equation}
\label{eq:grp-loss}
\ell^{\mathrm{grp}}(s,\thet)
=-\sum_{v\in\gV_{g}^{k}}q_{g}(v)\,\log p_{\thet}(v\mid\vct p),
\qquad
q_{g}(v)=\frac{\bar p_{g}(v)}{\sum_{u\in\gV_{g}^{k}}\bar p_{g}(u)}.
\end{equation}

Substituting $\ell^{\mathrm{grp}}$ into \eqref{eq:gradmm} replaces the class label with a soft group distribution read off the model's own knowledge, while preserving the separation between groups. Section~\ref{sec:ablation-signals} compares it against several alternatives.

\subsection{Online gradient matching with rehearsal memory}
\label{sec:method-online}

The objective of \eqref{eq:gradmm} suits an offline setup, where a single fixed dataset is available all at once. A deployed system rarely works this way, since its data arrives as a sequence of blocks $\Dreal^{1},\dots,\Dreal^{T}$, with $\Dreal^{1:t}$ the data seen through step $t$~\citep{li2025tic}. In this online setting, we want a compact synthetic set of at most $m$ examples that, at each step, best represents everything seen so far, so that fine-tuning $\thet_0$ on it tracks the model trained on all data through step $t$. We write this set as ${}^{\star}\Dsyn^{1:t}$, the one whose gradient at $\thet_0$ matches that of the full history:
\begin{equation}
\label{eq:online-ideal}
{}^{\star}\Dsyn^{1:t}=\argmin_{\substack{|\Dsyn^{1:t}|\le m,\ \Dsyn^{1:t}\subset\Gamma}}\;D\!\left(\nabla_{\thet}\,\ell(\Dsyn^{1:t},\thet_0),\ \nabla_{\thet}\,\ell(\Dreal^{1:t},\thet_0)\right).
\end{equation}

Computing ${}^{\star}\Dsyn^{1:t}$ directly is impractical in a stream, since it requires keeping the entire history $\Dreal^{1:t}$ and re-distilling it from scratch at every step, so both storage and per-step cost grow without bound as the stream lengthens. We approximate it by solving a separate distillation problem for each incoming block $\Dreal^{t}$ and merging the results into a running rehearsal memory through the same gradient-matching objective. Each block is first compressed on its own into a synthetic set $\Dsyn^{t}$ by the offline problem of \eqref{eq:gradmm}, and the memory $\gM^{t}$ carries these forward. It starts from the first distilled block and, at every later step, is rebuilt by distilling the union of the new set and the previous memory back down to $m$,
\begin{equation}
\label{eq:merge}
\gM^{t} = \argmin_{\substack{|\,\cdot\,|\le m,\ \,\cdot\,\subset\Gamma}}\;D\!\left(\nabla_{\thet}\,\ell(\,\cdot\,,\thet_0),\ \nabla_{\thet}\,\ell(\Dsyn^{t}\cup\gM^{t-1},\thet_0)\right),\qquad \gM^{1}=\Dsyn^{1}.
\end{equation}
We call this a \emph{cascaded merge} (Figure~\ref{fig:overview}(c)). At every step we fine-tune on $\gM^{t}$, which stands in for the ${}^{\star}\Dsyn^{1:t}$ we cannot compute. Each step distills only the new block and the current memory instead of the whole history, so its cost is the same at every step and the total grows linearly with the stream length $T$. Since the merge is independent of the loss, it applies to both the supervised and the label-free objectives. We compare the cascaded merge against alternative merge strategies in Section~\ref{sec:ablation-merge}.

\subsection{Hybrid hierarchical compression}
\label{sec:method-hybrid}

Selection and distillation work best in different regimes. When the budget is a large fraction of the data, cheap coreset selection is enough, since a well-chosen real subset still covers most of the training signal. As the budget shrinks toward a handful of examples, no real subset can carry that signal, whereas a single synthetic example can combine signal drawn from many real ones. Viewing compression as an information bottleneck~\citep{tishby2000information,zhou2025rome}, we expect selection to suffice while it is wide and distillation to be needed once it narrows.

We therefore compress in two stages (Figure~\ref{fig:overview}(d)). At every compression step, offline on a static pool or at each step of the stream, a coreset selector first cuts the incoming batch of $B$ examples down to a smaller pool of $K$, and gradient matching then distills that pool into the $m$ examples actually kept, so $B\gg K\gg m$. We call this the hybrid pipeline and compare it against selection-only and distillation-only alternatives in Section~\ref{sec:results-hybrid}.

\section{Experiments}
\label{sec:results}

\subsection{Experimental setup}
\label{sec:setup}

\paragraph{Data.} We use five text classification datasets: SST-2~\citep{socher2013recursive}, Rotten Tomatoes~\citep{pang2005seeing}, RT-Polarity~\citep{pang2005seeing}, IMDB~\citep{maas2011learning}, and TwitterEmotion~\citep{saravia2018carer}. They span two domains, movie reviews and tweets, inputs from a single sentence to a full review, and training pools from 1{,}000 examples to tens of thousands. Every experiment uses all five, except the streaming hybrid, whose ten steps of 1{,}000 examples only SST-2 and IMDB can supply. Appendix~\ref{sec:supp-data} details the data partitioning and stream construction.

\paragraph{Model.} Following GradMM~\citep{nguyen2025synthetic}, we use phi-1.5~\citep{li2023textbooks} for our experiments. During distillation, we only match the gradients of the output layer to avoid recomputing the gradients of all 1.3B parameters at each step, and we run the ADMM solver for 30 outer iterations of 50 inner steps at a step size of 0.008. During fine-tuning, we train 200 steps on a linear schedule, evaluating every 50 steps and keeping the best checkpoint, with the learning rate picked per dataset from $\{7{\times}10^{-6},\,1{\times}10^{-5},\,1.5{\times}10^{-5}\}$. While we use phi-1.5 for both distillation and fine-tuning, two different models can be used in general, without any change to the method, because what distillation returns is plain text rather than weights.

\paragraph{Compared methods and metrics.} In the offline setup, our main metric is accuracy after fine-tuning, and we compare against five competing methods: k-center greedy~\citep{sener2018active} takes the example farthest from those already picked; herding~\citep{rebuffi2017icarl} takes the ones that keep the running mean closest to the average; prototypical~\citep{snell2017prototypical} takes the ones closest to that average; uncertainty~\citep{lewis1994sequential} takes the ones the model is least sure about; and random picks uniformly. In the online setup, we focus on the area under the accuracy-versus-step curve (AUC), which rewards being right at every step rather than only the last, and we compare against four competing methods: k-center greedy, herding, and prototypical from above, and reservoir sampling~\citep{vitter1985random}, which keeps a uniform sample of all data seen. In both settings we also report three reference points. An oracle fine-tunes on all available real data, and supervised GradMM distills with the ground-truth labels that our label-free method never sees. Together they give upper bounds, while the zero-shot model with no fine-tuning gives the lower bound.

\subsection{Offline label-free gradient matching beats competing methods}
\label{sec:results-offline}

We benchmark OLGA, setting $k{=}100$ for its label-free group distribution objective, against the supervised objective (GradMM) and the five competing methods in the offline setup, where the full data pool is available at once. Each task starts from 50 real anchors, which we distill into a compact synthetic set of 5 examples and fine-tune phi-1.5 on, evaluating on the held-out validation split of each dataset. This is a 90\% reduction, which creates the extreme-compression regime we target.

\begin{figure}[t]
\centering
\includegraphics[width=\columnwidth]{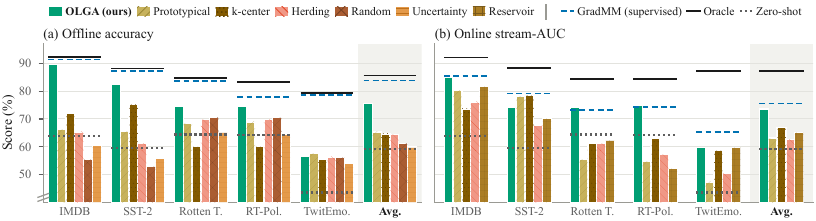}
\setlength{\abovecaptionskip}{-6.3pt}%
\caption{Across five datasets, OLGA beats every competing method on average in both the offline and online setups, and comes close to the supervised upper bound in the online one. \textbf{(a)} Offline methods reduce 50 real examples to 5, and we report the held-out accuracy of phi-1.5 fine-tuned on them. \textbf{(b)} Online data arrives over a sequence of steps, each method keeps a memory of only 5 examples, and we report stream-AUC, the area under the accuracy-versus-step curve.}
\label{fig:offline-online}
\vspace{-10pt}%
\end{figure}

Figure~\ref{fig:offline-online}(a) reports classification accuracy on each dataset. OLGA is the strongest label-free method, beating the strongest competing method on four of the five datasets and by 10.2 points on average. The gap is widest on IMDB, where OLGA scores 89.5 against 72.1 for the strongest competitor, and narrowest on Rotten Tomatoes, at 4.1 points (74.5 against 70.4). The one exception is TwitterEmotion, which is hard for every method, and where OLGA reaches just 56.5, on par with the competing methods. Overall, even without labels, gradient matching keeps enough signal to compress a dataset into a handful of synthetic examples that train a model better than any subset of real examples can. Table~\ref{tab:ex-main} shows two distilled examples next to the real examples they borrow their phrasing from, and Appendix~\ref{sec:supp-examples} one for each of the other datasets. Appendix~\ref{sec:supp-offline-values} lists exact values for Figure~\ref{fig:offline-online}(a).

\begin{table}[t]
\caption{Two synthetic examples distilled by OLGA, one per dataset. Each is generated as a mix of phrases from real examples, three of which are listed below it. A shared phrase has the same color in both places, and its superscript gives the number of the real example it comes from.}
\label{tab:ex-main}
\begin{center}\scriptsize
\setlength{\exlabw}{1.15cm}\setlength{\tabcolsep}{4pt}\renewcommand{\arraystretch}{1.05}
\begin{tabular}{@{}YW|W@{}}
\toprule
& \textbf{TwitterEmotion} (joy) & \textbf{SST-2} (negative) \\
\midrule
\exlab{Synthetic}
& \sharedcn{1}{I am feeling} \sharedcn{2}{so much} better after my recent health
surgery yesterday and \sharedcn{3}{I can feel} more
& teacher was giving \sharedcn{1}{a long} and \sharedcn{2}{boring and} not
\sharedcn{3}{interesting and} clear lecture to the students because the teacher was reading from a
textbook \\
\midrule
\exlab{Real}
& \exitem{1}{\sharedc{1}{i am feeling} amazing mostly normal i am going to a pre thanksgiving celebration with our friends from that time we were in softball}%
\exitem{2}{i feel like we have \sharedc{2}{so much} to be thankful for}%
\exitem{3}{\sharedc{3}{i can feel} superior on that point}%
& \exitem{1}{it ai n't art, by \sharedc{1}{a long} shot,}%
\exitem{2}{a shame that stealing harvard is too busy getting in its own way to be anything but frustrating, \sharedc{2}{boring, and} forgettable.}%
\exitem{3}{a more confused, less \sharedc{3}{interesting and} more sloppily made film could possibly come down the road in 2002}%
\\
\bottomrule
\end{tabular}
\end{center}
\vspace{-15.5pt}%
\end{table}

\subsection{Online gradient matching with memory beats continual selection}
\label{sec:results-stream}

We benchmark OLGA, setting $m{=}5$ for the rehearsal memory and cascaded merge budgets, across all five datasets in an online setup of $T{=}10$ steps ($8$ for IMDB). At each step, 50 anchors arrive and are distilled into 5 synthetic examples, which are then folded into the memory by the cascaded merge. At each step, phi-1.5 is fine-tuned on the resulting memory and evaluated on the held-out validation set.

Figure~\ref{fig:offline-online}(b) reports the online stream-AUC of the fine-tuned phi-1.5 on each dataset. Averaged over the five datasets, Online GradMM and OLGA finish 8.5 and 6.6 points above k-center, the strongest competing method overall at 67.0. The two are within two points of each other, so removing the labels costs little. Across datasets the biggest gain for both comes on RT-Polarity, roughly 10 points over the best competing method there. The shortfalls are all on the label-free side, where OLGA only draws level with the best competing method on TwitterEmotion and comes in 4.2 points behind it on SST-2. Appendix~\ref{sec:supp-online-values} lists exact values for Figure~\ref{fig:offline-online}(b).

\subsection{Hybrid compression pipeline}
\label{sec:results-hybrid}

We explore whether a mix of selection and distillation exists that lets cheap selection take the early, easy part of compression and leaves only the extreme end to distillation, and how such a mix compares against alternatives. In our setup, a batch of $B{=}1{,}000$ real examples arrives, selection trims it to $K{=}50$, and distillation turns those into the $m{=}5$ examples we keep and train on. This achieves a 99.5\% compression rate overall. We call this two-stage route \textbf{Hybrid}, and compare it against two alternatives. \textbf{Direct} drops the middle stage and goes from $B$ straight to $m$ in a single step. \textbf{Double} keeps both stages but runs the second one the same way as the first. Across all three routes, the distillation stage runs either GradMM or OLGA, and selection is done by uniform sampling.

\paragraph{Hybrid pipeline in offline setting.} Table~\ref{tab:hybrid} compares the accuracy of the three routes in the offline setting, where the whole compression happens once over a static pool, and Table~\ref{tab:hybrid-runtime} compares what each route costs in time. With labels, the hybrid pipeline is the strongest route, averaging 86.0 percent across the five datasets, ahead of GradMM-Direct at 82.7 and GradMM-Double at 73.9, while costing less than distilling the whole batch directly or in two steps. This confirms that selecting early, while the compression is still easy, and distilling only at the extreme end is the more effective way through the bottleneck, and the cheaper one too. Without labels, the same funnel becomes a trade-off rather than a gain: OLGA-Hybrid falls short of OLGA-Direct by only 0.9 points on average while running 30\% faster, which puts the two routes on par and makes the choice between them a balance of accuracy against time. What holds for both objectives is the lead over selection: GradMM-Hybrid and OLGA-Hybrid beat the selection-only routes on average, 86.0 and 80.0 against 73.4 for Selection-Double and 66.6 for Selection-Direct.

\begin{table}[t]
\caption{Comparison of the hybrid, direct, and double routes in the offline and online settings across 5 datasets. At each time step, the data is compressed in stages from $B{=}1{,}000$ to $K{=}50$ to $m{=}5$. The offline setting runs a single step ($T{=}1$), and the online setting runs $T{=}10$ steps. Runtime is the wall-clock time of one offline distillation run on SST-2. On average, the hybrid route beats the direct and double routes for both objectives in both settings, except offline without labels, where OLGA-Hybrid is on par with OLGA-Direct. Bold marks the best result in each column, and underline marks the reference that each colored difference is measured against.}
\label{tab:hybrid}
\label{tab:hybrid-runtime}
\begin{center}
\resizebox{\columnwidth}{!}{%
\begin{tabular}{ll|ccccc|l|cc|l|l}
\toprule
\multirow{2}{*}{\textbf{Method}} & \multirow{2}{*}{\textbf{Route}} & \multicolumn{6}{c|}{\textbf{Offline accuracy}} & \multicolumn{3}{c|}{\textbf{Online stream-AUC}} & \multirow{2}{*}{\textbf{Runtime (s) $\downarrow$}} \\
\cmidrule(lr){3-8} \cmidrule(lr){9-11}
 & & \textbf{SST-2 $\uparrow$} & \textbf{Rotten T. $\uparrow$} & \textbf{IMDB $\uparrow$} & \textbf{RT-Pol. $\uparrow$} & \textbf{TwitEmo. $\uparrow$} & \textbf{Avg. $\uparrow$} & \textbf{SST-2 $\uparrow$} & \textbf{IMDB $\uparrow$} & \textbf{Avg. $\uparrow$} & \\
\midrule
\multirow{3}{*}{GradMM} & Hybrid & 88.2 & 83.9 & \textbf{92.3} & 82.1 & \textbf{83.4} & \textbf{86.0}\dgain{3.3} & \textbf{76.0} & \textbf{80.8} & \textbf{78.4}\dgain{3.1} & 436.1\dfaster{29.9\%} \\
 & Direct & 87.4 & 76.6 & 90.6 & 81.3 & 77.6 & \underline{82.7} & 72.1 & 78.5 & \underline{75.3} & \underline{622.1} \\
 & Double & 79.7 & 71.1 & 80.7 & 67.0 & 70.8 & 73.9\dloss{8.8} & --- & --- & --- & 2073.6\dslower{233.3\%} \\
\midrule
\multirow{3}{*}{OLGA} & Hybrid & 87.8 & 81.2 & 89.0 & \textbf{84.6} & 57.3 & 80.0\dloss{0.9} & 71.2 & 78.4 & 74.8\dgain{2.1} & 447.5\dfaster{30.1\%} \\
 & Direct & \textbf{88.6} & \textbf{85.0} & 90.9 & 82.6 & 57.4 & \underline{80.9} & 67.7 & 77.7 & \underline{72.7} & \underline{639.9} \\
 & Double & 67.4 & 71.2 & 87.6 & 76.1 & 58.9 & 72.2\dloss{8.7} & --- & --- & --- & 2086.4\dslower{226.1\%} \\
\midrule
\multirow{2}{*}{Selection} & Double & 82.5 & 73.6 & 89.8 & 52.9 & 68.0 & 73.4\dgain{6.8} & 69.6 & 75.3 & 72.5\dgain{1.4} & 5.5\dslower{14.6\%} \\
 & Direct & 81.9 & 54.7 & 87.9 & 50.3 & 58.1 & \underline{66.6} & 67.2 & 75.0 & \underline{71.1} & \underline{\textbf{4.8}} \\
\bottomrule
\end{tabular}%
}
\end{center}
\vspace{-10.5pt}%
\end{table}

\paragraph{Hybrid pipeline in online setting.} We extend the comparison to the online setting, where a new batch of $B$ examples arrives at each of $T{=}10$ steps. After each distillation, the resulting $m$ examples are merged into the running rehearsal memory by the cascaded merge. We report only SST-2 and IMDB, since they are the only datasets large enough to supply a stream of this length.

The online half of Table~\ref{tab:hybrid} shows that GradMM-Hybrid performs best, with an average stream-AUC of 78.4. GradMM-Hybrid beats GradMM-Direct by 3.1 points and OLGA-Hybrid beats OLGA-Direct by 2.1 points, while distilling only 50 examples at each step instead of 1{,}000. Every route that uses distillation also beats selection alone on average, and the best of them leads the strongest selection-only route (72.5) by 5.9 points.

\subsection{Ablations}
\label{sec:ablations}

\begin{figure}[t]
\begin{center}
\includegraphics[width=\columnwidth]{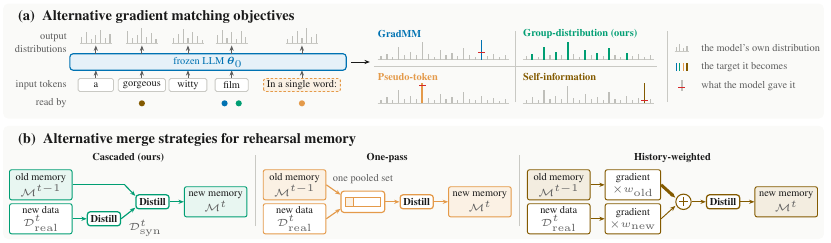}
\end{center}
\caption{\textbf{(a)} How each gradient matching objective
builds its loss from the model's next-token distribution; the dots mark which
position it reads. \textbf{(b)} Three ways to fold a new batch into the memory:
compress it first, pool it with the memory, or weight the two by hand.}
\label{fig:ablation-overview}
\vspace{-9pt}%
\end{figure}

\paragraph{Alternative label-free objectives.}
\label{sec:ablation-signals}
The group-distribution loss $\ell^{\mathrm{grp}}$ introduced in OLGA is not the only way to turn the model's own probabilities into a label-free signal. We compare it against two alternatives (Figure~\ref{fig:ablation-overview}(a)). The \emph{pseudo-token} signal appends the prompt ``\,In a single word:'' and takes the most likely next token as its target, letting the model label itself in one word. The \emph{self-information} signal finds the token in each sequence that the frozen model predicted least well and takes it as the target, on the view that this token carries most of what makes the example distinctive. Appendix~\ref{sec:supp-signals} describes both signals further and gives their exact formulations.

\begin{figure}[t]
\centering
\includegraphics[width=\columnwidth]{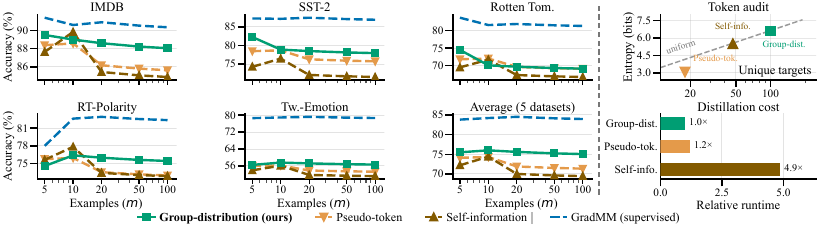}
\caption{Comparison of alternative label-free objectives. \textbf{(Left)} Accuracy against the number of examples $m$, per dataset and on average. \textbf{(Right)} Number of distinct tokens each signal aims at, and runtime relative to group-distribution. Group-distribution leads at $m{=}5$, stays steady at larger budgets, and runs fastest.}
\label{fig:ablation-signals}
\vspace{-9pt}%
\end{figure}

Figure~\ref{fig:ablation-signals} compares the three signals. At $m{=}5$, group-distribution averages 75.5 percent across the five datasets, ahead of pseudo-token (74.1) and self-information (72.2), and it is the steadiest as the budget grows. The tokens each signal aims at help explain why. Pseudo-token aims at only 18 distinct tokens, spread over just 3.1 bits of entropy, so its training signal keeps pushing in the same few directions, while self-information spreads over 47 tokens (5.5 bits) and group-distribution over 100 (6.6 bits). The alternatives are also up to five times slower to run.

\paragraph{Alternative selectors for the hybrid pipeline.}
\label{sec:ablation-selector}
The hybrid routes use uniform sampling by default, but any selection method can take its place. We repeat the offline GradMM-Hybrid route with k-center, the strongest selection baseline of Figure~\ref{fig:offline-online}(b). On average, uniform sampling is both more accurate and faster than k-center (Table~\ref{tab:ablation-selector}). This indicates that naive sampling, which keeps the original distribution of the data, is preferable in the hybrid pipeline, since distillation is able to synthesize higher-quality text from it. Appendix~\ref{sec:supp-selector} reports the per-dataset numbers.

\begin{table}[t]
\begin{minipage}[t]{0.49\textwidth}
\caption{Effect of the selector on the offline hybrid pipeline, averaged over the five datasets.}
\label{tab:ablation-selector}
\end{minipage}\hfill
\begin{minipage}[t]{0.49\textwidth}
\caption{Comparison of online memory-merging strategies, averaged over the five datasets.}
\label{tab:ablation-merge}
\end{minipage}

\vspace{6pt}
\begin{minipage}[t]{0.49\textwidth}
\centering\fontsize{8}{9.5}\selectfont
\setlength{\tabcolsep}{3.7pt}
\begin{tabular}[t]{lcc}
\toprule
\textbf{Selector} & \textbf{Avg. accuracy $\uparrow$} & \textbf{SST-2 runtime (s) $\downarrow$} \\
\midrule
\textbf{Uniform (ours)} & \textbf{86.0} & \textbf{436.1} \\
k-center & 82.1 & 564.8 \\
& & \\  %
\bottomrule
\end{tabular}
\end{minipage}\hfill
\begin{minipage}[t]{0.49\textwidth}
\centering\fontsize{8}{9.5}\selectfont
\setlength{\tabcolsep}{4pt}
\begin{tabular}[t]{lc}
\toprule
\textbf{Merge strategy} & \textbf{Avg. stream-AUC $\uparrow$} \\
\midrule
\textbf{Cascaded (ours)} & \textbf{72.0} \\
One-pass & 70.2 \\
History-weighted & 69.8 \\
\bottomrule
\end{tabular}
\end{minipage}
\vspace{-4pt}%
\end{table}

\paragraph{Alternative merge strategies.}
\label{sec:ablation-merge}
The cascaded merge is only one way to use distillation to fold an arriving batch into the running memory. We compare it against two alternatives (Figure~\ref{fig:ablation-overview}(b)). The \emph{one-pass} merge appends the old memory to the raw new batch and distills the two together in a single step. The \emph{history-weighted} merge does the same single pass, but weights the old and new gradients by how much data each one stands for.

Table~\ref{tab:ablation-merge} extends the experiment of Section~\ref{sec:results-stream} for OLGA with the alternative merge strategies, at $T{=}5$ steps. The cascaded merge beats both on average, at 72.0 against 70.2 for one-pass and 69.8 for history-weighted. Appendix~\ref{sec:supp-merge} reports the per-dataset performance.

\section{Conclusion}
\label{sec:conclusion}

Keeping a deployed language model current calls for updates that run on a handful of examples carrying the training signal of all data seen so far. OLGA extends gradient matching for text in three directions: removing the need for labels, following a data stream through a running rehearsal memory, and cutting cost with a hybrid pipeline that selects before it distills. Across five text classification datasets, it outperforms every competing method on average in both the offline and online settings, and our experiments show that each of the three directions matters. Together, they make continual updating of a language model accurate, affordable, and independent of ground-truth labels, offering a practical path toward keeping deployed language model applications current.

\section*{AI Use Statement}

The research ideas and hypotheses, the method and its formulation in Section~\ref{sec:method}, the design of the experiments, the implementation of the methods, and the analysis and interpretation of the results are all our own work. Generative AI tools were used only in a supporting role. AI coding agents also wrote secondary engineering code, namely experiment scripts, job launchers, and the plotting code behind our figures, and large language models were used to edit the paper for readability. No generative AI tools were used to generate, clean, or reformat datasets, for translation, or for qualitative data analysis. The synthetic examples reported in the paper are the output of our method run with phi-1.5, not text produced by an AI assistant. We have reviewed all AI-assisted work. We reviewed every piece of code the agents produced. We take responsibility for the final content of this work, including text, claims or artifacts produced with the aid of generative AI.

\section*{Reproducibility Statement}

All experiments use publicly available datasets (SST-2, Rotten Tomatoes, RT-Polarity, IMDB, and TwitterEmotion) and the open-source phi-1.5 model. We will release our full implementation publicly on GitHub upon acceptance, and it reproduces all results reported in the paper.

\section*{Ethics Statement}

We have read and adhere to the ICLR Code of Ethics. Our experiments use only publicly available text classification datasets and an open-source language model, and involve no human subjects and no data collection of our own. Because the synthetic examples our method produces are built from phrases of the underlying data, they may carry over biases present in that data, which should be considered before deploying the method.

\bibliography{iclr2027_conference}
\bibliographystyle{iclr2027_conference}

\clearpage
\appendix
\renewcommand{\thesection}{A\arabic{section}}
\section*{Appendix}
\addcontentsline{toc}{section}{Appendix}

\section{Data partitioning and construction of data streams}
\label{sec:supp-data}

Each dataset is partitioned into a training split and a validation split. SST-2, Rotten Tomatoes, and TwitterEmotion come with these splits by default, and we use them as provided, while IMDB and RT-Polarity do not, so we construct a fixed split for them once. Within each validation split, we hold out 100 examples, chosen with a fixed seed, as a source pool, and the remaining examples form the evaluation set, which serves every method, step, and experiment, and on which no experiment trains. The offline experiments draw their real anchors from the source pool, and the offline hybrid, which needs a much larger batch, draws it from the training split together with the source pool.

The online experiments stream the training split instead. We fix one random permutation of it and feed it in consecutive disjoint blocks, one block per step. The online hybrid streams the full official training split in the same way but balanced across groups, so each step receives a fresh batch of $B{=}1{,}000$ examples. Only SST-2 and IMDB have training splits large enough for this.

Budgets $m$, $K$, and $B$ are counted per group throughout the paper. All five tasks are binary, so each stage actually handles twice the stated number of examples; the offline hybrid batch, for instance, holds $B{=}1{,}000$ examples of each group. Since the budgets before and after each compression stage double together, the compression rates are unchanged, and for simplicity we leave the number of groups out of the notation.

\section{Exact values for the offline results}
\label{sec:supp-offline-values}

Bar charts make the offline comparison easy to scan but hard to read exactly, so here we list the values behind Section~\ref{sec:results-offline} in full. Each method reduces 50 real examples to 5, phi-1.5 is fine-tuned on the result, and Table~\ref{tab:offline-perdataset} reports its accuracy on the held-out validation split of each dataset, together with its gap to the best selection method.

\begin{table}[H]
\caption{Exact values for Figure~\ref{fig:offline-online}(a), the offline accuracy of phi-1.5 on each dataset. The upper block gives the supervised upper bounds and the lower block the label-free methods. OLGA beats the best selection method on four of the five datasets and by 10.2 points on average. Bold marks the best label-free result in each column, and underline marks the reference that each colored difference is measured against.}
\label{tab:offline-perdataset}
\begin{center}
\resizebox{\columnwidth}{!}{%
\begin{tabular}{llllll|l}
\toprule
\textbf{Method} & \textbf{IMDB $\uparrow$} & \textbf{SST-2 $\uparrow$} & \textbf{Rotten Tom. $\uparrow$} & \textbf{RT-Polarity $\uparrow$} & \textbf{TwitEmo. $\uparrow$} & \textbf{Avg. $\uparrow$} \\
\midrule
Oracle & 92.3\dgain{20.2} & 88.2\dgain{12.9} & 84.7\dgain{14.3} & 83.2\dgain{12.8} & 79.4\dgain{21.9} & 85.6\dgain{20.3} \\
GradMM~\citep{nguyen2025synthetic} & 91.4\dgain{19.3} & 87.3\dgain{12.0} & 83.7\dgain{13.3} & 78.0\dgain{7.6} & 78.6\dgain{21.1} & 83.8\dgain{18.5} \\ 
\midrule
\textbf{OLGA (ours)} & \textbf{89.5}\dgain{17.4} & \textbf{82.3}\dgain{7.0} & \textbf{74.5}\dgain{4.1} & \textbf{74.6}\dgain{4.2} & 56.5\dloss{1.0} & \textbf{75.5}\dgain{10.2} \\ 
Prototypical~\citep{snell2017prototypical} & 66.3\dloss{5.8} & 65.4\dloss{9.9} & 68.5\dloss{1.9} & 68.9\dloss{1.5} & \underline{\textbf{57.5}} & \underline{65.3} \\ 
k-center~\citep{sener2018active} & \underline{72.1} & \underline{75.3} & 60.1\dloss{10.3} & 60.2\dloss{10.2} & 55.4\dloss{2.1} & 64.6\dloss{0.7} \\ 
Herding~\citep{rebuffi2017icarl} & 65.3\dloss{6.8} & 61.2\dloss{14.1} & 69.8\dloss{0.6} & 69.7\dloss{0.7} & 56.3\dloss{1.2} & 64.5\dloss{0.8} \\ 
Random & 55.4\dloss{16.7} & 52.9\dloss{22.4} & \underline{70.4} & \underline{70.4} & 56.2\dloss{1.3} & 61.1\dloss{4.2} \\ 
Uncertainty~\citep{lewis1994sequential} & 60.4\dloss{11.7} & 55.8\dloss{19.5} & 65.0\dloss{5.4} & 64.5\dloss{5.9} & 54.0\dloss{3.5} & 59.9\dloss{5.4} \\ 
Zero-shot & 64.0\dloss{8.1} & 59.4\dloss{15.9} & 64.4\dloss{6.0} & 64.3\dloss{6.1} & 43.5\dloss{14.0} & 59.1\dloss{6.2} \\
\bottomrule
\end{tabular}%
}
\end{center}
\end{table}

\section{Exact values for the online results}
\label{sec:supp-online-values}

As with the offline results, we list the exact values behind the online comparison of Section~\ref{sec:results-stream} here. Data arrives over $T{=}10$ steps ($8$ for IMDB), each method keeps a memory of 5 examples, and Table~\ref{tab:stream} reports the resulting stream-AUC on each dataset, together with each entry's gap to the best competing method.

\begin{table}[H]
\caption{Exact values for Figure~\ref{fig:offline-online}(b), the online stream-AUC of phi-1.5 on each dataset. Oracle-accumulate keeps all data seen so far and gives the upper bound. OLGA beats the best competing method by 6.6 points on average and comes within two points of supervised GradMM. Bold marks the best result in each column apart from the oracle, and underline marks the reference that each colored difference is measured against.}
\label{tab:stream}
\begin{center}
\resizebox{\columnwidth}{!}{%
\begin{tabular}{llllll|l}
\toprule
\textbf{Method} & \textbf{IMDB $\uparrow$} & \textbf{SST-2 $\uparrow$} & \textbf{TwitEmo. $\uparrow$} & \textbf{Rotten Tom. $\uparrow$} & \textbf{RT-Polarity $\uparrow$} & \textbf{Avg. $\uparrow$} \\
\midrule
Oracle-accumulate & 92.1\dgain{4.8} & 88.4\dgain{10.0} & 87.2\dgain{27.5} & 84.5\dgain{20.1} & 84.3\dgain{20.0} & 87.3\dgain{18.2} \\
\midrule
\textbf{Online GradMM (ours)} & \textbf{85.4}\dgain{3.8} & \textbf{79.2}\dgain{0.8} & \textbf{65.3}\dgain{5.6} & 73.2\dgain{8.8} & 74.3\dgain{10.0} & \textbf{75.5}\dgain{8.5} \\ 
\textbf{OLGA (ours)} & 85.1\dgain{3.5} & 74.2\dloss{4.2} & 59.6\dloss{0.1} & \textbf{74.1}\dgain{9.7} & \textbf{74.9}\dgain{10.6} & 73.6\dgain{6.6} \\ 
\midrule
k-center~\citep{sener2018active} & 73.6\dloss{8.0} & \underline{78.4} & 58.8\dloss{0.9} & 61.3\dloss{3.1} & 62.9\dloss{1.4} & \underline{67.0} \\ 
Reservoir~\citep{vitter1985random} & \underline{81.6} & 70.3\dloss{8.1} & \underline{59.7} & 62.1\dloss{2.3} & 52.3\dloss{12.0} & 65.2\dloss{1.8} \\ 
Herding~\citep{rebuffi2017icarl} & 76.1\dloss{5.5} & 67.7\dloss{10.7} & 50.5\dloss{9.2} & 61.3\dloss{3.1} & 57.1\dloss{7.2} & 62.5\dloss{4.5} \\ 
Prototypical~\citep{snell2017prototypical} & 80.4\dloss{1.2} & 78.0\dloss{0.4} & 47.3\dloss{12.4} & 55.3\dloss{9.1} & 54.7\dloss{9.6} & 63.1\dloss{3.9} \\ 
Zero-shot & 64.0\dloss{17.6} & 59.4\dloss{19.0} & 43.5\dloss{16.2} & \underline{64.4} & \underline{64.3} & 59.1\dloss{7.9} \\
\bottomrule
\end{tabular}%
}
\end{center}
\end{table}

\section{Extending the offline results to $m{=}10$ examples}
\label{sec:supp-perdataset-b10}

Distillation is designed for extreme compression, so it is worth checking how the picture changes when the budget is less tight. We repeat the offline comparison of Section~\ref{sec:results-offline} at $m{=}10$ examples, with the same methods and setup as Table~\ref{tab:offline-perdataset}. Table~\ref{tab:supp-offline-perdataset-b10} gives the per-dataset results. The conclusions of Section~\ref{sec:results-offline} carry over, as OLGA remains the strongest label-free method on average, 1.7 points ahead of the best selection method, and supervised GradMM stays close to the oracle. The margin is narrower than at $m{=}5$, however, since selection improves faster as the budget grows and already overtakes OLGA on three of the five datasets at $m{=}10$. Distillation is strongest under extreme compression while selection catches up as the budget grows, which is the split the hybrid pipeline is built on and is consistent with its results in Table~\ref{tab:hybrid}.

\begin{table}[h]
\caption{Offline accuracy of phi-1.5 on each dataset at $m{=}10$, the counterpart to Table~\ref{tab:offline-perdataset}. The upper block gives the supervised upper bounds and the lower block the label-free methods. OLGA stays ahead of the best selection method by 1.7 points on average, but leads on only two of the five datasets. Bold marks the best label-free result in each column, and underline marks the reference that each colored difference is measured against.}
\label{tab:supp-offline-perdataset-b10}
\begin{center}
\resizebox{\columnwidth}{!}{%
\begin{tabular}{llllll|l}
\toprule
\textbf{Method} & \textbf{IMDB $\uparrow$} & \textbf{SST-2 $\uparrow$} & \textbf{Rotten Tom. $\uparrow$} & \textbf{RT-Polarity $\uparrow$} & \textbf{TwitEmo. $\uparrow$} & \textbf{Avg. $\uparrow$} \\
\midrule
Oracle & 92.3\dgain{11.9} & 88.2\dgain{6.2} & 84.7\dgain{10.2} & 83.2\dgain{9.1} & 79.4\dgain{16.9} & 85.6\dgain{12.9} \\
GradMM~\citep{nguyen2025synthetic} & 90.6\dgain{10.2} & 87.2\dgain{5.2} & 81.6\dgain{7.1} & 82.6\dgain{8.5} & 78.9\dgain{16.4} & 84.2\dgain{11.5} \\ 
\midrule
\textbf{OLGA (ours)} & \textbf{89.0}\dgain{8.6} & 78.9\dloss{3.1} & 70.1\dloss{4.4} & \textbf{76.4}\dgain{2.3} & 57.6\dloss{4.9} & \textbf{74.4}\dgain{1.7} \\ 
Prototypical~\citep{snell2017prototypical} & 78.8\dloss{1.6} & 80.1\dloss{1.9} & 72.2\dloss{2.3} & 72.0\dloss{2.1} & 60.2\dloss{2.3} & \underline{72.7} \\ 
Herding~\citep{rebuffi2017icarl} & 77.6\dloss{2.8} & 78.8\dloss{3.2} & 71.5\dloss{3.0} & 71.8\dloss{2.3} & 58.4\dloss{4.1} & 71.6\dloss{1.1} \\ 
Uncertainty~\citep{lewis1994sequential} & 70.4\dloss{10.0} & 70.5\dloss{11.5} & 73.0\dloss{1.5} & 72.8\dloss{1.3} & \underline{\textbf{62.5}} & 69.8\dloss{2.9} \\ 
k-center~\citep{sener2018active} & \underline{80.4} & \underline{\textbf{82.0}} & 61.1\dloss{13.4} & 61.6\dloss{12.5} & 61.0\dloss{1.5} & 69.2\dloss{3.5} \\ 
Random & 68.3\dloss{12.1} & 66.1\dloss{15.9} & \underline{\textbf{74.5}} & \underline{\textbf{74.1}} & 62.1\dloss{0.4} & 69.0\dloss{3.7} \\ 
Zero-shot & 64.0\dloss{16.4} & 59.4\dloss{22.6} & 64.4\dloss{10.1} & 64.3\dloss{9.8} & 43.5\dloss{19.0} & 59.1\dloss{13.6} \\
\bottomrule
\end{tabular}%
}
\end{center}
\end{table}

\section{What a distilled example looks like}
\label{sec:supp-examples}

Because OLGA returns plain text rather than weights, its synthetic examples can be read directly. OLGA optimizes each synthetic example so that its training gradient matches that of many real examples at once, and then maps it back to real tokens. The text it settles on therefore reads as a blend of those real examples, taking a phrase from one and another phrase from a different one, and joining them into a sentence that appears nowhere in the data. Tables~\ref{tab:ex-imdb} to~\ref{tab:ex-rtpol} show one such example for each of the remaining datasets, in the style of Table~\ref{tab:ex-main}. In each table, the synthetic example comes first and the real examples it draws from are listed below it. A phrase shared between the two is highlighted in both places, and its superscript gives the number of the real example it comes from.

\setlength{\intextsep}{10pt plus 2pt minus 2pt}

\begin{table}[htbp]
\caption{Synthetic example distilled by OLGA for \textbf{IMDB} (positive), with the seven real examples it borrows its phrasing from. A shared phrase is highlighted in both places, and its superscript gives the number of the real example it comes from.}
\label{tab:ex-imdb}
\centering\footnotesize
\setlength{\tabcolsep}{3pt}\renewcommand{\arraystretch}{1.1}
\begin{tabular}{@{}YZ@{}}
\toprule
\exlab{Synthetic} & IMDLEG \sharedn{1}{is great} because it is a \sharedn{2}{well made} and entertaining
story that \sharedn{3}{keeps you} engaged \sharedn{4}{from beginning} onwards and the well done
production and the talented \sharedn{5}{and well} done \sharedn{6}{people who} are involved in
making the show are truly \sharedn{7}{a great and} a \\
\midrule
\exlab{Real} & \exitem{1}{This is just a short comment but I stumbled onto this movie by chance and I loved it. The acting \shared{is great}, the story is simple and touching \ldots\ Seek it out.}%
\exitem{2}{I thought this was one of those really great films to see with a bunch of close friends. \ldots\ Although not a new concept this was a very \shared{well made} film.}%
\exitem{3}{Powers Boothe turns in a stellar performance as 1970's cult figure Jim Jones of the Peoples Temple. \ldots\ The movie \shared{keeps you} riveted and is a must see for anyone.}%
\exitem{4}{A funny comedy \shared{from beginning} to end! \ldots\ Creative story line with a very talented cast. I thoroughly enjoyed it!}%
\exitem{5}{terribly underrated with matt dillon and tom skerritt, good backdrop for solid story \ldots\ well acted \shared{and well} cast, tommy lee jones and bruce dern make you hate them with passion}%
\exitem{6}{This movie will always be a Broadway and Movie classic, as long as there are still \shared{people who} sing, dance, and act.}%
\exitem{7}{\shared{A great and} truly independent film that hit most of my emotions and carried me into another world. \ldots}%
\\
\bottomrule
\end{tabular}
\end{table}

\begin{table}[htbp]
\caption{Synthetic example distilled by OLGA for \textbf{Rotten Tomatoes} (negative), with the four real examples it borrows its phrasing from. A shared phrase is highlighted in both places, and its superscript gives the number of the real example it comes from.}
\label{tab:ex-rt}
\centering\footnotesize
\setlength{\tabcolsep}{3pt}\renewcommand{\arraystretch}{1.1}
\begin{tabular}{@{}YZ@{}}
\toprule
\exlab{Synthetic} & The new movie was boring and had no \sharedn{1}{interesting or} well developed
action or \sharedn{2}{a good} one \sharedn{3}{of the actors} in \sharedn{4}{the story} \\
\midrule
\exlab{Real} & \exitem{1}{what the director can't do is make either of val kilmer's two personas \shared{interesting or} worth caring about.}%
\exitem{2}{the techno tux is good for a few laughs, as are chan and hewitt, but when such \shared{a good} design turns out to be a cheap knockoff \ldots}%
\exitem{3}{sadly, though many \shared{of the actors} throw off a spark or two when they first appear, they can't generate enough heat in this cold vacuum of a comedy \ldots}%
\exitem{4}{\shared{the story} drifts so inexorably into cliches about tortured (and torturing) artists \ldots}%
\\
\bottomrule
\end{tabular}
\end{table}

\begin{table}[htbp]
\caption{Synthetic example distilled by OLGA for \textbf{RT-Polarity} (positive), with the three real examples it borrows its phrasing from. A shared phrase is highlighted in both places, and its superscript gives the number of the real example it comes from.}
\label{tab:ex-rtpol}
\centering\footnotesize
\setlength{\tabcolsep}{3pt}\renewcommand{\arraystretch}{1.1}
\begin{tabular}{@{}YZ@{}}
\toprule
\exlab{Synthetic} & Amazing movie review provided by critics \sharedn{1}{is a great}
\sharedn{2}{example of how} a single positive sentence can \sharedn{3}{make a} significant impact
on the \\
\midrule
\exlab{Real} & \exitem{1}{morton \shared{is a great} actress portraying a complex character, but morvern callar grows less compelling the farther it meanders from its shocking start.}%
\exitem{2}{a truly moving experience, and a perfect \shared{example of how} art -- when done right -- can help heal, clarify, and comfort.}%
\exitem{3}{australian actor/director john polson and award-winning english cinematographer giles nuttgens \shared{make a} terrific effort at disguising the obvious with energy and innovation.}%
\\
\bottomrule
\end{tabular}
\end{table}

\setlength{\intextsep}{12pt plus 2pt minus 2pt}

\FloatBarrier
\section{Alternative label-free gradient signals}
\label{sec:supp-signals}

The group-distribution loss \eqref{eq:grp-loss} is one of several ways to read a matching target off the frozen model. Here we give the exact definitions of the two alternatives compared against it in Section~\ref{sec:ablation-signals}, pseudo-token and self-information. Both are built like $\ell^{\mathrm{sup}}$ and $\ell^{\mathrm{grp}}$, as a per-example cross-entropy toward a target the data supplies. Each substitutes into \eqref{eq:gradmm} in place of $\ell$, and the same loss forms both the real and the synthetic gradient.

\paragraph{Pseudo-token.} Let $\vct u$ be the fixed suffix ``\,In a single word:'' and write $\vct p\oplus\vct u$ for the prompt with that suffix appended. The loss is the cross-entropy toward the token the frozen model finds most likely to come next,
\[
\ell^{\mathrm{pt}}(s,\thet)=-\log p_{\thet}\!\left(\hat v(s) \mid \vct p\oplus\vct u\right),
\qquad
\hat v(s)=\argmax_{v}\ p_{\thet_0}\!\left(v \mid \vct p\oplus\vct u\right).
\]
The model is asked to name the example in one word and then trained to keep saying that word. Like $\ell^{\mathrm{sup}}$ this puts all the weight on a single token, but the token is whatever the model volunteers rather than a class name we supply.

\paragraph{Self-information.} No suffix is added here. Writing $s=(\vct p,\vct r)$ with response tokens $r_1,\dots,r_n$, the quantity $-\log p_{\thet_0}(r_i \mid \vct p,r_{<i})$ says how unexpected the frozen model found the token at position $i$. The loss is the cross-entropy at the position where it is largest,
\[
\ell^{\mathrm{si}}(s,\thet)=-\log p_{\thet}\!\left(r_{i^{\star}} \mid \vct p,r_{<i^{\star}}\right),
\qquad
i^{\star}(s)=\argmax_{1\le i\le n}\ \bigl[-\log p_{\thet_0}\!\left(r_i \mid \vct p,r_{<i}\right)\bigr].
\]
The reasoning is that the token the model predicts least well is the one that carries most of what makes the example distinctive.

Two differences from \eqref{eq:grp-loss} follow from these definitions. Both targets above are single tokens, where the group-distribution target spreads over the $k$ tokens of $\gV_{g}^{k}$; and both depend on the individual example, where the group-distribution target depends only on the group it belongs to. For the real anchors neither difference costs anything, since every target is computed once under $\thet_0$ before matching starts. For the synthetic set, the second one does, because the synthetic sequences keep changing while the solver runs, so a target read off a sequence has to be found again as it changes, whereas the group-distribution target is fixed from the outset. This is what separates the three runtimes in Figure~\ref{fig:ablation-signals}, and self-information is the most expensive of them because finding its target means scoring every position of every sequence inside the matching loop.

\section{Alternative selectors for the hybrid pipeline}
\label{sec:supp-selector}

The first stage of the hybrid pipeline only has to thin out the incoming batch, so any selection method can run it. As in Section~\ref{sec:ablation-selector}, we replace uniform sampling in the first stage of the offline GradMM-Hybrid route with k-center, the strongest selection method of Figure~\ref{fig:offline-online}(b), and keep everything else fixed. Table~\ref{tab:supp-selector-full} gives the per-dataset results. Uniform sampling is ahead of k-center on all five datasets, from 2.0 points on IMDB to 8.3 on TwitterEmotion, and is also faster to run. This supports keeping the original distribution of the data at the selection stage and leaving the rest of the compression to distillation.

\begin{table}[H]
\caption{Per-dataset version of Table~\ref{tab:ablation-selector}, the offline accuracy of the GradMM-Hybrid route with uniform sampling or k-center as its selector. Uniform sampling is more accurate on every dataset and faster to run. Bold marks the better selector in each column.}
\label{tab:supp-selector-full}
\begin{center}
\resizebox{\columnwidth}{!}{%
\begin{tabular}{lccccc|c|c}
\toprule
\textbf{Selector} & \textbf{SST-2 $\uparrow$} & \textbf{Rotten Tomatoes $\uparrow$} & \textbf{IMDB $\uparrow$} & \textbf{RT-Polarity $\uparrow$} & \textbf{TwitterEmotion $\uparrow$} & \textbf{Avg. $\uparrow$} & \textbf{SST-2 runtime (s) $\downarrow$} \\
\midrule
\textbf{Uniform (ours)} & \textbf{88.2} & \textbf{83.9} & \textbf{92.3} & \textbf{82.1} & \textbf{83.4} & \textbf{86.0} & \textbf{436.1} \\
k-center & 84.2 & 80.9 & 90.3 & 79.8 & 75.1 & 82.1 & 564.8 \\
\bottomrule
\end{tabular}%
}
\end{center}
\end{table}

\section{Alternative merge strategies}
\label{sec:supp-merge}

The running memory can absorb each arriving batch in more than one way. As in Section~\ref{sec:ablation-merge}, we compare the cascaded merge against two alternatives for folding each batch into the memory of OLGA, the one-pass and the history-weighted merge, over $T{=}5$ steps. Table~\ref{tab:supp-merge-full} gives the per-dataset stream-AUC. The cascaded merge scores highest on SST-2, Rotten Tomatoes, and RT-Polarity, while one-pass leads on TwitterEmotion and history-weighted on IMDB. Each alternative wins on one dataset at the cost of the others, which leaves the cascaded merge ahead on average, at 72.0 against 70.2 and 69.8.

\begin{table}[H]
\caption{Per-dataset version of Table~\ref{tab:ablation-merge}, the online stream-AUC of OLGA under each merge strategy. The cascaded merge leads on three of the five datasets and on average. Bold marks the best result in each column.}
\label{tab:supp-merge-full}
\begin{center}
\resizebox{\columnwidth}{!}{%
\begin{tabular}{lccccc|c}
\toprule
\textbf{Merge strategy} & \textbf{SST-2 $\uparrow$} & \textbf{Rotten Tomatoes $\uparrow$} & \textbf{IMDB $\uparrow$} & \textbf{RT-Polarity $\uparrow$} & \textbf{TwitterEmotion $\uparrow$} & \textbf{Avg. $\uparrow$} \\
\midrule
\textbf{Cascaded (ours)} & \textbf{61.8} & \textbf{85.4} & 69.1 & \textbf{69.5} & 73.9 & \textbf{72.0} \\
One-pass & 60.4 & 78.7 & 69.3 & 62.5 & \textbf{80.1} & 70.2 \\
History-weighted & 57.2 & 85.3 & \textbf{72.4} & 68.1 & 66.0 & 69.8 \\
\bottomrule
\end{tabular}%
}
\end{center}
\end{table}

\section{Memory learnability diagnostics}
\label{sec:supp-learnability}

\begin{figure}[htbp]
\centering
\includegraphics[width=\columnwidth]{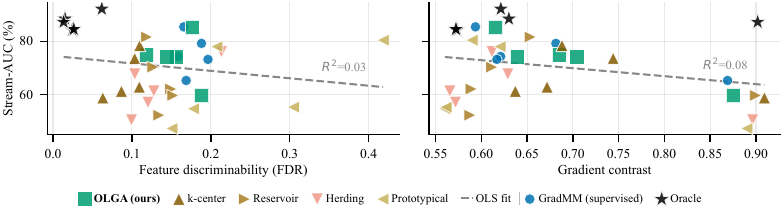}
\setlength{\abovecaptionskip}{-6.4pt}%
\caption{Feature discriminability (left) and gradient contrast (right) of each memory against its stream-AUC, one point per dataset and method. Neither property predicts how well a memory trains the model.}
\label{fig:supp-learnability}
\end{figure}

This section asks whether a cheap property of the memory can tell in advance how well the model will do on the stream, without fine-tuning on it. We measure two such properties of each method's memory on every dataset. Feature discriminability measures how well the memory's examples separate the classes in the model's feature space, and gradient contrast measures how different the training gradients of the classes are. Figure~\ref{fig:supp-learnability} plots each against stream-AUC. Neither one predicts stream-AUC on its own. The least-squares fits explain almost none of the variation, with $R^2$ of 0.03 for feature discriminability and 0.08 for gradient contrast, and memories with nearly the same score on either property end up tens of points apart in stream-AUC. The gradient-matching memories and the oracle do sit in the high-accuracy region, but so do some selection memories whose scores are very different from theirs. How well a memory trains the model therefore cannot be read off a single cheap property, and the choice of method, rather than a proxy score, has to drive how the memory is built.

\end{document}

%% file: math_commands.tex
\usepackage{amsmath,amsfonts,bm}

\def\eqref#1{equation~\ref{#1}}

\def\1{\bm{1}}

\DeclareMathAlphabet{\mathsfit}{\encodingdefault}{\sfdefault}{m}{sl}
\SetMathAlphabet{\mathsfit}{bold}{\encodingdefault}{\sfdefault}{bx}{n}

\def\gD{{\mathcal{D}}}

\def\gM{{\mathcal{M}}}

\def\gV{{\mathcal{V}}}

\DeclareMathOperator*{\argmax}{arg\,max}
\DeclareMathOperator*{\argmin}{arg\,min}